\documentclass[a4paper, 10pt, conference]{ieeeconf}      
\usepackage{graphicx}
\usepackage{cite}
\usepackage{hyperref}
\usepackage{textgreek}
\usepackage{amsmath}
\usepackage{amsfonts}
\usepackage{mathtools}
\usepackage{xcolor}
\usepackage{amssymb}
\usepackage{float}
\usepackage{subcaption}
\usepackage{latexsym}
\usepackage{grffile}
\newtheorem{theorem}{Theorem}
\newtheorem{lemma}{Lemma}

\IEEEoverridecommandlockouts                              

\title{\LARGE \bf
Concentration bounds for SSP Q-learning for average cost MDPs
}

\author{Shaan ul Haque$^{1}$ and Vivek S.\ Borkar$^{2,*}$
\thanks{*Work of VSB was  supported by an S.\ S.\ Bhatnagar Fellowship from Council for Scientific and Industrial Research, Govt.\ of India}
\thanks{$^{1}$Department of Electrical Engineering, IIT Bombay, Powai, Mumbai 400076, India
        {\tt\small shaanhaque2016@gmail.com}}%
\thanks{$^{2}$Department of Electrical Engineering, IIT Bombay, Powai, Mumbai 400076, India
        {\tt\small borkar.vs@gmail.com}}%
}

\begin{document}

\maketitle
\thispagestyle{empty}
\pagestyle{empty}

\begin{abstract}
We derive a concentration bound for a Q-learning algorithm for average cost Markov decision processes based on an equivalent shortest path problem, and compare it numerically with the alternative scheme based on relative value iteration.

\end{abstract}

\section{Introduction}
Q-learning, introduced originally for the discounted cost Markov decision processes in \cite{Watkins},  is a data-driven reinforcement learning
algorithm for learning the `Q-factor' function arising from the dynamic programming
equation for the infinite horizon discounted reward problem. It can be viewed as a stochastic approximation counterpart of the classical value iteration for computing the value function arising as the solution of the corresponding dynamic programming equation. Going over from value function to the so called Q-factors facilitates an interchange of the conditional expectation and the nonlinearity (the minimization, to be precise) in the recursion, making it amenable for stochastic approximation. These ideas, however, do not extend automatically to the average cost problem, which is harder to analyze even when the model (i.e., the controlled transition probabilities) is readily available. The reason for this is the non-contractive nature of the associated Bellman operator.  This extension was achieved in \cite{abounadi} in two different ways. The first, called RVI Q-learning, is a stochastic approximation counterpart of the `relative value iteration' (or RVI) algorithm for average cost \cite{Puterman} and is close in spirit to the original. However, there is another algorithm dubbed SSP Q-learning based on an alternative scheme due to Bertsekas \cite{bert}, which does involve a contraction under a weighted max-norm. Motivated by a recent paper on concentration for stochastic approximation in \cite{chandak}, we present here a similar concentration for the SSP Q-learning exploiting its explicitly contractive nature, something that is missing in RVI, leading to  non-trivial technical issues in providing finite time guarantees for it (see, e.g., \cite{Zhang}). We also provide an empirical  comparison between the two with suggestive outcomes.

Section II builds up the background and section III states the key assumptions and the main result. Its proof follows in section IV. Section V describes the numerical experiments.

\section{Background}
\subsection{Preliminaries}
We consider a controlled Markov chain $\{X_n\}$ on a finite state space $S = \{1,2,...,d\}$ with a finite action space $U = \{u_1,...,u_r\}$ and transition
probabilities $p(j|i,u) =$ the probability of transition from $i$ to $j$ under action $u$ for $i, j \in S, u \in U$. Associated with this transition is a “running cost” $k: S\times U \mapsto \mathcal{R}$ and the aim is to choose actions $\{Z_n\}$ non-anticipatively (i.e., conditionally independent of the future state trajectory given past states and actions) so as to minimize the “average cost”
\begin{equation}\label{cost}
    \limsup_{n\to\infty}\frac{1}{n}\sum_{m=0}^{n-1}E[k(X_m,Z_m)].
\end{equation}
We shall be interested in “stationary policies” wherein $Z_n=v(X_n)$ for a map
$v : S \to U$. It is known that an optimal stationary policy exists under the following “unichain” condition which we assume throughout: under any stationary policy the chain has a single communicating class containing a common state (say, $i_0$). The dynamic programming equation for the above  problem is \cite{Puterman}
\begin{equation}\label{avg_v}
    V(i)=\min_u\Big[k(i,u)+\sum_{j\in \mathcal{S}}p(j|i,u)V(j)-\beta\Big].
\end{equation}
The unknowns are $V(\cdot),\beta$ where  $\beta$ is uniquely characterized as the optimal  average cost. $V(\cdot)$ is only unique upto an additive constant. The associated ``Q-factor" is
\begin{equation}
    Q(i,a)=\Big[k(i,u)+\sum_{j\in \mathcal{S}}p(j|i,u)\min_vQ(j,v)-\beta\Big].
\end{equation}
The aim is to get these Q-factors even when we do not know the transition probabilities, but have access to a black box which can generate random variables according to the above transition probabilities. 
\subsection{SSP Q-learning}\label{sec:ssp}
Recall the stochastic shortest path problem. Let  $S = S_0\cup T$ with $S_0\cap T=\phi$ and $i_0 \in T$. The objective is to minimize
$$E\Big[\sum_{n=1}^{\tau-1}k(X_n,Z_n)+h(X_{\tau})\Big]$$ where $h: S\times U \mapsto \mathcal{R}$ is the terminal cost and $\tau := \min\{n\geq 0:X_n\in T\}$. Under our assumtions, $\tau < \infty$ a.s., in fact $E[\tau] < \infty$. The dynamic programming equation to solve this problem is given by
\begin{gather*}
    V(i)=\min_u\big[k(i,u)+\sum_{j\in S}p(j|i,u)V(j)\big]] \ \forall \ i\in S_0,\\
    V(i)=h(i) \ \forall \ i\in T.
\end{gather*}
Coming back to average cost problem, SSP Q-learning is based on the observation that the average cost under any stationary policy is simply the ratio of the expected total cost and the expected time, between two successive visits to the reference state $i_0$. This connection was exploited by \cite{bert} to convert the average cost problem into a stochastic shortest path (SSP)  problem. Consider a family of SSP problem parameterized by $\lambda$, with the cost given by $k(i,u)-\lambda$ for $k$ as above and some scalar $\lambda$. 
Then the dynamic programming equation for the above SSP problem is
\begin{subequations}
    \begin{gather}\label{ssp}
    V(i)=\min_u\big[k(i,u)+\sum_{j\in S, j\neq i_0}p(j|i,u)V(j)-\lambda \big],\\
    V(i_0)=0.
\end{gather}
\end{subequations}
For each fixed policy, the cost is linear in $\lambda$ with negative slope. Thus $V(\cdot)$, being the lower envelope thereof, is piecewise linear with finitely many linear pieces and concave decreasing in $\lambda$ for each component. When we replace $\lambda$ by $\beta$ and force $V(i_0)=0$, we recover  (\ref{avg_v}). This suggests the coupled iterations
\begin{subequations}
\begin{gather}\label{ssp_vi}
    V_{k+1}(i)=\min_u\big[k(i,u)+\sum_{j\in S, j\neq i_0}p(j|i,u)V_k(j)-\lambda_k \big],\\
    \lambda_{k+1}=\lambda_k+a(n)V_k(i_0).
\end{gather}
\end{subequations}
The SSP Q-learning scheme for the above problem is \cite{abounadi}
\begin{subequations}
\begin{gather}
    \nonumber Q_{n+1}(i,u)=Q_n(i,u)+a(n)I\{X_n=i,Z_n=u\}\big(k(i,u)\\
+ \ \min_vQ_n(X_{n+1},v)I\{X_{n+1}\neq i_0\}-\lambda_n-Q_n(i,u)\big),\\
    \lambda_{n+1}=\Gamma(\lambda_n+a'(n)\min_vQ_n(i_0,v)).
\end{gather}
\end{subequations}
Here $\Gamma(\cdot)$ is a projection operator onto the interval $[-g,g]$ with $g$ chosen so as to satisfy $\beta\in(-g,g)$. Although this assumes some prior knowledge of $\beta$, that can be obtained by a bound on $k(\cdot,\cdot)$. This also ensures that (\ref{F_tilde}) below holds. We rewrite the above equations as follows
\begin{subequations}
\begin{gather}
    \nonumber Q_{n+1}(i,u)=Q_n(i,u)+a(n)[F^{i,u}(Q_n, (X_n,Z_n), \lambda_n)\\ -Q_n(i,u)+M^{i,u}_{n+1}(Q_n)],\\
    \lambda_{n+1}=\Gamma(\lambda_n+a'(n)f(Q_n)),
\end{gather}
\end{subequations}
and
\begin{gather*}
    F^{i,u}(Q_n, (X_n,Z_n), \lambda_n)=I\{X_n=i, Z_n=u\}\big(k(i,u)\\
    +\sum_{j\neq i_0} p(j|i,u)\min_vQ_n(j,v)-\lambda_n\big),\\
    M^{i,u}_{n+1}(Q_n)=I\{X_n=i, Z_n=u\}\big(\min_vQ_n(X_{n+1},v)\times\\ I\{X_{n+1}\neq i_0\}-\sum_{j\neq i_0} p(j|i,u)\min_vQ_n(j,v)\big),\\
f(Q) = \min_uQ(i_0,u).
\end{gather*}
As observed in \cite{tsitsiklis}, the map $F(\cdot,\cdot,\lambda)$ is a contraction for a fixed $\lambda$ under a certain weighted max-norm
$$\|x\|:=\max|w_ix_i|, \ x\in \mathcal{R}^d,$$ for an appropriate weight vector $w = [w_1,...,w_d]$, $w_i > 0$.

\section{Main Result}
We state our main theorem in this section, after setting up the notation and assumptions. The assumptions are specifically geared for the SSP Q-learning applications in Section \ref{sec:ssp}, as will become apparent. 

Consider the coupled iteration 
\begin{gather}\label{iteration}
    x_{n+1}=x_n+a(n)\big(F(x_n,Y_n,\lambda_n)-x_n+M_{n+1}(x_n)\big),\\
    \lambda_{n+1}=\Gamma(\lambda_n+a'(n)f(x_n)), n\geq 0
\end{gather}
for $x_n=[x_n(1),...,x_n(d)]^T\in\mathcal{R}^d, \lambda_n\in\mathcal{R}$. Here:
\begin{itemize}
    \item $\{Y_n\}$ is the `Markov noise' taking values in a finite state space $S$, i.e.,
    \begin{eqnarray*}
        P(Y_{n+1}|Y_m,x_m, m\leq n)&=&P(Y_{n+1}|Y_n,x_n)\\
        &=&p_{x_n}(Y_{n+1}|Y_n), \ n\geq 0,
    \end{eqnarray*}
where for each $w$, $p_w(\cdot|\cdot)$ is the transition probability of an irreducible Markov chain on $S$ with unique stationary distribution $\pi_w$. We assume that the map $w\mapsto p_w(j|i)$ is Lipschitz, i.e., for some $L_1>0$,
$$\sum_{j\in S} |p_w(j|i)-p_v(j|i)|\leq L_1 \|w-v\|,\ \forall i\in S, w, v\in\mathcal{R}^d.$$
By Cramer's theorem, $\pi_w(i)$ is a rational function of $\{p_w(j|k)\}$ with a non-vaishing denominator, so the map $w\mapsto \pi_w(i)$ is similarly Lipschitz, i.e., for some $L_2>0$,
 $$\sum_{i\in S}|\pi_w(i)-\pi_v(i)|\leq L_2 \|w-v\|,\ \forall i\in S, w, v\in\mathcal{R}^d.$$
 See  Appendix B, \cite{chandak} for some bounds on $L_2$. 

    \item $\{M_n(x)\}$ is, for each $x\in\mathcal{R}^d$, an $\mathcal{R}^d$-valued martingale difference sequence parametrized by $x$, with respect to the increasing family of $\sigma$-fields $\mathcal{F}_n\coloneqq\sigma(x_0, Y_m, M_m(x), x\in\mathcal{R}^d,m\leq n)$, $n\geq0$. That is,
    \begin{equation}\label{MDS}
        E[M_{n+1}(x)|\mathcal{F}_n]=\theta \: \ \mathrm{a.s.} \: \ \forall \; x,n,
    \end{equation}
 where $\theta$ is the zero vector.   We also assume the  componentwise bound: for some $K_0>0$,
    \begin{equation}\label{MDS_assumption}
        |M_n^\ell(x)|\leq K_0(1+\|x\|) \:\textrm{a.s.} \: \forall \; x,n,l.
    \end{equation}
    
    \item 
    $F(\cdot) = [F^1(\cdot), \cdots , F^d(\cdot)]^T:\mathcal{R}^d\times S\rightarrow\mathcal{R}^d$ satisfies
    \begin{gather}\label{Contraction}
        \nonumber \|\sum_{i\in S}\pi_w(i)(F(x,i,\lambda)-F(z,i,\lambda))\|\leq \alpha \|x-z\|,\\ \ \forall \ x,z,w\in\mathcal{R}^d, \lambda\in\mathcal{R},
    \end{gather}
    for some $\alpha\in(0,1)$. By the contraction mapping theorem, this implies that $\sum_i\pi_w(i)F(\cdot,i, \lambda)$ has a unique fixed point $x^*_w(\lambda)\in\mathcal{R}^d$ (i.e., $\sum_i\pi_w(i)F(x^*_w(\lambda),i,\lambda)$ $=x^*_w(\lambda)$). We assume that $x^*_w(\lambda)$ is independent of $w$, i.e., there exists a $x^*(\cdot)$ such that
    \begin{equation}\label{fixed_point}
        \sum_{i\in S}\pi_w(i)F(x^*(\lambda),i, \lambda)=x^*(\lambda), \\ \ \forall \;w\in\mathcal{R}^d.
    \end{equation}
     We also assume that the map $x\mapsto F^\ell(x,i,\lambda)$ is Lipschitz (w.l.o.g., uniformly in $i$ and $\ell$). Let the common Lipschitz constant be $L_3>0$, i.e., 
    $$|F^\ell(x,i,\lambda)-F^\ell(z,i,\lambda)|\leq L_3\|x-z\|,$$ $$\ \forall i\in S, \ell\in\{1,\cdots, d\}, x,z\in\mathcal{R}^d.$$ 
We assume that  $F$ is concave piecewise linear and decreasing in $\lambda$. Furthermore, $\widetilde{F}_n(x,Y_n,\lambda)\coloneqq F(x,Y_n,\lambda)+M_{n+1}(x)$ is assumed to satisfy
    \begin{gather}\label{F_tilde}
        \|\widetilde{F}_n(x,Y_n,\lambda)\|\leq K+\alpha\|x\| \: \ \textrm{a.s.}.
    \end{gather}
    
    \item Moreover, we assume that $f(x^*(\lambda))$ is Lipschitz with Lipchitz constant $L_4>0$:    for all $\lambda_1, \lambda_2\in \mathcal{R}$
    \begin{equation}
         |f(x^*(\lambda_1))-f(x^*(\lambda_2))|\leq L_4||x^*(\lambda_1)-x^*(\lambda_2)||.
    \end{equation}

    \item $a(n) \geq 0$ is a sequence of stepsizes satisfying 
    \begin{equation}
     a(n) \to 0, \    \sum_na(n)=\infty,
    \end{equation}
and is assumed to be eventually non-increasing, i.e., there exists $n^*\geq1$ such that $a(n+1)\leq a(n) \ \forall \; n\geq n^*$. Since $a(n)\to 0$, there exists $n^\dagger$ such that $a(n)<1$ for all $n>n^\dagger$.\footnote{Observe that we do not require the classical square-summability condition in stochastic approximation, viz., $\sum_na(n)^2 < \infty$. This is because the contractive nature of our iterates gives us an additional handle on errors by putting less weight on past errors. A similar effect was observed in \cite{chandak}.} 
We further assume that $a(n)=\Omega(1/n)$. So, $a(n)\geq \frac{d_1}{n}$ for all $n\geq n_1$ for some $n_1$ and $d_1>0$. We also assume that there exists $0<d_2 \leq1$ such that $a(n)=\mathcal{O}\left(\left(\frac{1}{n}\right)^{d_2}\right)$, i.e., $a(n)\leq d_3\left(\frac{1}{n}\right)^{d_2}$ for all $n\geq n_2$ for some $n_2$ and $d_3>0$. Larger values of $d_1$ and $d_2$ and smaller values of $d_3$ improve the main result presented below. The role this assumption plays in our bounds will become clear later. Define $N\coloneqq \max\{n^*, n^\dagger, n_1, n_2\}$, i.e., $a(N)<1$, $a(n)$ is non-increasing after $N$ and $\frac{d_1}{n}\leq a(n)\leq d_3\left(\frac{1}{n}\right)^{d_2}, \forall\ n\geq N$. Also, it is assumed that the sequence $\epsilon(n)=\frac{a'(n)}{a(n)} \to 0,$ i.e., $a'(n)=o(a(n))$.
\end{itemize}

\ \ \ For $n\geq0$, we further define:
\begin{eqnarray*}
    b_k(n) &=& \sum_{m=k}^na(m), \ 0\leq k\leq n<\infty, \\
    b'_k(n) &=& \sum_{m=k}^na'(m), \ 0\leq k\leq n<\infty, \\
   \color{black}\beta_k(n) &\color{black}=& \color{black}\begin{cases}
      \frac{1}{k^{d_2-d_1}n^{d_1}}, & \text{if}\ d_1\leq d_2 \\
      \frac{1}{n^{d_2}}, & \text{otherwise},
    \end{cases} \\
    \kappa(d) &=& \|\textbf{1}\|, \ \textbf{1} := [1,1,\cdots , 1]^T \in \mathcal{R}^d, \ d \geq 1,\\
\bar{\epsilon}(n) &=& \sup_{m\geq n}\epsilon(m).
\end{eqnarray*}

Our main result is a follows: \\

\begin{theorem}\label{shaan_th}
    
    (a)  Let $n_0\geq N$. Then there exist finite positive constants $c_1$, $c_2$, $c_3$ and $D$, depending on $\|x_N\|$, such that for $\delta>0$, $n\geq n_0$ and $C=e^{\kappa(d)(K_0(1+\|x_N\|+\frac{K}{1-\alpha})+c_2)}$, the inequality
    \begin{gather}\label{inequality_a}
        \nonumber \|x_n-x^*(\lambda_n)\|\leq e^{-(1-\alpha)b_{n_0}(n)}\|x_{n_0}-x^*(\lambda_{n_0})\|\\+\frac{\delta+a(n_0)c_1+\bar{\epsilon}(n_0)c_3}{1-\alpha}
    \end{gather}
    holds with probability exceeding
    \begin{gather}
    1-2d\sum_{m=n_0+1}^ne^{-D\delta^2/\beta_{n_0}(m)}, 0<\delta\leq C,\\
    1-2d\sum_{m=n_0+1}^ne^{-D\delta/\beta_{n_0}(m)}, \delta> C
    \end{gather}
    
    (b) There exist finite constants $c_4$, $c_5$ and an $N'\geq N$ large enough such that for $n\geq \hat{n}\geq N'$, the inequality
    \begin{gather}\label{inequality_b}
        \nonumber |\lambda_n-\beta|\leq |\lambda_{\hat{n}}-\beta|e^{-(1-\alpha)c_4b'_{\hat{n}}(n)}\\+
        c_5\Big(e^{-(1-\alpha)b_{\hat{n}}(n)}\|x_{\hat{n}}-x^*(\lambda_{\hat{n}})\|+\frac{\delta+a(\hat{n})c_1+\bar{\epsilon}(\hat{n})c_3}{1-\alpha}\Big)
    \end{gather}
    holds with probability exceeding
    \begin{gather}
        1-2d\sum_{m\geq\hat{n}+1}^ne^{-D\delta^2/\beta_{\hat{n}}(m)}, 0<\delta\leq C,\\
        1-2d\sum_{m\geq\hat{n}+1}^ne^{-D\delta/\beta_{\hat{n}}(m)}, \delta> C
    \end{gather}
    
\end{theorem}

\section{Proof}
We begin  with a  lemma adapted from \cite{chandak}.
\begin{lemma}\label{Bound_x_n}
    $\sup_{n\geq N}\|x_n\|\leq\|x_N\|+\frac{K}{1-\alpha},$ a.s.
\end{lemma}
    \proof
    Using (\ref{F_tilde}), we have
    \begin{eqnarray*}
    \|x_{n+1}\|&=&\|(1-a(n))x_n+a(n)\widetilde{F}_n(x_n,Y_n,\lambda_n)\| \nonumber\\
    &\leq&(1-a(n))\|x_n\|+a(n)(K+\alpha\|x_n\|) \nonumber\\
    &=&(1-(1-\alpha)a(n))\|x_n\|+a(n)K.
    \end{eqnarray*}
    For $n,m\geq N$, define $\psi(n,m)\coloneqq\prod_{k=m}^{n-1}(1-(1-\alpha)a(k))$ if $n>m$ and $1$ otherwise. Note that, since $a(N)<1$, $0\leq\psi(n,m)\leq1$ for all $n,m\geq N$. Then
    \begin{equation*}
    \|x_{n+1}\|-\frac{K}{1-\alpha}\leq (1-(1-\alpha)a(n))\left(\|x_n\|-\frac{K}{1-\alpha}\right).
    \end{equation*}
    Now $\|x_{N}\|\leq\|x_{N}\|+\frac{K}{1-\alpha}=\psi(N,N)\|x_{K}\|+\frac{K}{1-\alpha}$. Suppose
    \begin{equation}\label{induction_step}
        \|x_n\|\leq\psi(n,N)\|x_{N}\|+\frac{K}{1-\alpha}
    \end{equation}
    for some $n\geq N$. Then,
    \begin{eqnarray*}
    \|x_{n+1}\|-\frac{K}{1-\alpha}&\leq&(1-(1-\alpha)a(n))\Big(\psi(n,N)\|x_{N}\|\\&+&\frac{K}{1-\alpha}-\frac{K}{1-\alpha}\Big)\nonumber\\
    &\leq&\psi(n+1,N)\|x_{N}\|
    \end{eqnarray*}
    By induction, (\ref{induction_step}) holds for all $n\geq N$, which completes the proof of Lemma \ref{Bound_x_n}. 
    \endproof
    
\subsection{Concentration bound for the first iteration} 
Define $z_{n_0}=x_{n_0}$ and for $n>n_0$:
\begin{gather}\label{det_eq}
    z_{n+1}=z_n+a(n)\big(\sum_{i\in S} \pi_{x_n}(i)F(z_n, i, \lambda_n)-z_n\big),\\ 
    \lambda_{n+1}=\Gamma(\lambda_n+a'(n)f(x_n)), n\geq 0.
\end{gather}

We use the following theorem adapted from \cite{chandak}, which gives a concentration inequality for the stochastic approximation algorithm with Markov noise.\\

\begin{theorem}\label{chandak_th}
    Let $n_0\geq N$. Then there exist finite  constants $c_1, c_2, D > 0$, depending on $\|x_N\|$, such that for $\delta>0$, $n\geq n_0$ and $C=e^{\kappa(d)(K_0(1+\|x_N\|+\frac{K}{1-\alpha})+c_2)}$,
the inequality
\begin{gather}
    \|x_n-z_n\|\leq \frac{\delta+a(n_0)c_1}{1-\alpha}, \ n \geq n_0,
\end{gather}
    holds with probability exceeding
    \begin{gather}
    1-2d\sum_{m=n_0+1}^ne^{-D\delta^2/\beta_{n_0}(m)}, 0<\delta\leq C,\\
    1-2d\sum_{m=n_0+1}^ne^{-D\delta/\beta_{n_0}(m)}, \delta> C
    \end{gather}
\end{theorem}
\medskip

Since $F(x^*(\lambda_n), i, \lambda_n))=x^*(\lambda_n)$, we have
\begin{gather}
    \nonumber z_{n+1}-x^*(\lambda_{n+1})=(1-a(n))(z_n-x^*(\lambda_n))\\ +\nonumber a(n)\big(\sum_{i\in S} \pi_{x_n}(i)(F(z_n, i, \lambda_n)-F(x^*(\lambda_n), i, \lambda_n)) \\+\frac{1}{a(n)}(x^*(\lambda_n)-x^*(\lambda_{n+1}))\big).
\end{gather}
Since the map $\lambda \to F$ is piecewise linear and concave decreasing, and therefore so is the map $\lambda \to x^*(\lambda)$. By (\ref{Contraction}) and (\ref{fixed_point}), we have the following lemma,
\begin{lemma}
    $||x^*(\lambda_{n+1})-x^*(\lambda_n)|| \leq L_x|\lambda_{n+1}-\lambda_n|$
\end{lemma}

\proof From the definition of $x^*(\cdot)$, we have
\begin{gather*}
    \|x^*(\lambda_{n+1})-x^*(\lambda_n)\|= \|\sum_{i\in S}\pi(i)\big(F(x^*(\lambda_n), i, \lambda_n)\\-F(x^*(\lambda_{n+1}), i, \lambda_{n+1})\big)\|
\end{gather*}
We have suppressed the subscript of $\pi$, which irrelevant  by virtue of (\ref{fixed_point}). Let $e_i\in \mathcal{R}^d$ denote the standard basis vectors. Then the r.h.s.\ in the above can be written as
\begin{gather*}
     \|\sum_{i\in S}\pi(i)\big(F(x^*(\lambda_n), i, \lambda_n)-F(x^*(\lambda_{n+1}), i, \lambda_{n+1})\big)\|\\= \|\sum_{i\in S}\pi(i)\big(F(x^*(\lambda_n), i, \lambda_{n+1})-F(x^*(\lambda_{n+1}), i, \lambda_{n+1})\\+e_i(\lambda_{n+1}-\lambda_n)\big)\|\\ \leq \|\sum_{i\in S}\pi(i)\big(F(x^*(\lambda_n), i, \lambda_{n+1})-F(x^*(\lambda_{n+1}), i, \lambda_{n+1})\big)\|\\+\|\sum_{i\in S}\pi(i)\big(e_i(\lambda_{n+1}-\lambda_n)\big)\|\\
     \leq \alpha\|x^*(\lambda_{n+1})-x^*(\lambda_n)\|+|\lambda_{n+1}-\lambda_n|\|\sum_{i\in S}\pi(i)e_i\|.
\end{gather*}
Thus we finally have
\begin{gather*}
    \|x^*(\lambda_{n+1})-x^*(\lambda_n)\|\leq \alpha\|x^*(\lambda_{n+1})-x^*(\lambda_n)\| \\+|\lambda_{n+1}-\lambda_n|\|\sum_{i\in S}\pi(i)e_i\|
\end{gather*}
which leads us to the claim that
\begin{gather*}
    \|x^*(\lambda_{n+1})-x^*(\lambda_n)\|\leq L_x|\lambda_{n+1}-\lambda_n|
\end{gather*}
where $L_x=\|\pi\|/(1-\alpha)$. \endproof

To get a bound on $|\lambda_{n+1}-\lambda_n|$ we use the nonexpansive property of the projection operator as follows
\begin{gather*}
   |\lambda_{n+1}-\lambda_n|=|\Gamma(\lambda_n+a'(n)f(x_n))-\Gamma(\lambda_n)|\\
   \leq |\lambda_n+a'(n)f(x_n)-\lambda_n|= a'(n)|f(x_n)|
\end{gather*}
where we use the fact that $\Gamma(\lambda_n)=\lambda_n$. Combining the above inequalities, we get
\begin{equation}\label{x^*_bound}
    \|x^*(\lambda_{n+1})-x^*(\lambda_n)\|\leq L_xa'(n)|f(x_n)|.
\end{equation}
Thus,
\begin{gather}
    \|z_{n+1}-x^*(\lambda_{n+1})\|\leq \nonumber
    \|(1-a(n))(z_n-x^*(\lambda_n))\|\\ +\nonumber a(n)\big(\|\sum_{i\in S} \pi_{x_n}(i)(F(z_n, i, \lambda_n)-F(x^*(\lambda_n), i, \lambda_n))\| \\ \nonumber +\frac{1}{a(n)}\|(x^*(\lambda_n)-x^*(\lambda_{n+1}))\|\big)\\
    \leq (1-(1-\alpha)a(n))\|z_n-x^*(\lambda_n)\|+a(n)\epsilon(n)L_x|f(x_n)|\label{det_bound}.
\end{gather} 
where $\epsilon(n)=\frac{a'(n)}{a(n)}$. Since $x_n$ is bounded by Lemma \ref{Bound_x_n},  $f(x_n)\leq K^*<\infty$. 
Iterating (\ref{det_bound}) for $n_0\leq m\leq n$, we get,
\begin{gather}
    \nonumber \|z_{n+1}-x^*(\lambda_{n+1})\|\leq \\ \nonumber \|x_{n_0}-x^*(\lambda_{n_0})\|\prod_{m=n_0}^n(1-(1-\alpha)a(m))\\ +K'\sum_{m=n_0}^n\psi(n+1,m+1)a(m)\epsilon(m)\\
    \nonumber \leq e^{-(1-\alpha)b_{n_0}(n)}\|x_{n_0}-x^*(\lambda_{n_0})\| \\+K'\sum_{m=n_0}^n\psi(n+1,m+1)a(m)\epsilon(m)
\end{gather}
where, $K'=L_xK^*$ and $z_{n_0}=x_{n_0}$. 
The summation in last term can be bounded as 
\begin{gather}\label{psi_sum}
    \nonumber \sum_{m=n_0}^n\psi(n+1,m+1)a(m)\epsilon(m)\\ \leq \bar{\epsilon}(n_0)\sum_{m=n_0}^n\psi(n+1,m+1)a(m),
\end{gather}
where $\bar{\epsilon}(n) := \sup_{m\geq n}\epsilon(m)$. Note that for any $0<k< m$,
$$\psi(m,k)+\psi(m,k+1)(1-\alpha)a(k)=\psi(m,k+1),$$ and hence
$$\frac{\psi(m+1,n_0)}{1-\alpha}+\frac{1}{1-\alpha}\sum_{k={n_0}}^m\psi(m+1,k+1)(1-\alpha)a(k)=$$ $$\frac{\psi(m+1,m+1)}{1-\alpha}=\frac{1}{1-\alpha}.$$ This implies that 
\begin{equation}\label{psi_bound}
    \sum_{k=n_0}^m\psi(m+1,k+1)a(k) \ \leq \ \frac{1}{1-\alpha}.
\end{equation}
Hence
\begin{equation}
     \sum_{m=n_0}^n\psi(n+1,m+1)a(m)\bar{\epsilon}(m) \ \leq \ \frac{\bar{\epsilon}(n_0)}{1-\alpha}.
\end{equation}
Combining  the above,
\begin{eqnarray}
    \|z_{n+1}-x^*(\lambda_{n+1})\| &\leq& e^{-(1-\alpha)b_{n_0}(n)}\|x_{n_0}-x^*(\lambda_{n_0})\| \nonumber \\
&&+ \ \frac{K'\bar{\epsilon}(n_0)}{1-\alpha}. \label{det_conc}
\end{eqnarray}
Combining (\ref{det_conc}) with Theorem \ref{chandak_th} yields Theorem \ref{shaan_th}(a). 

\subsection{Concentration bound for the second iteration}
The second iteration is given by
\begin{equation}\label{sec_ite}
    \lambda_{n+1}=\Gamma(\lambda_n+a'(n)f(x)).
\end{equation}

Let $\xi_n=f(x_n)-f(x^*(\lambda_n))$. Subtracting $\beta$ from both sides, we get:
\begin{gather}
 \lambda_{n+1}-\beta= \Gamma(\lambda_n+a'(n)f(x^*(\lambda_n))+\nonumber\\  a'(n)(f(x_n)-f(x^*(\lambda_n)))\big)-\beta \nonumber\\  
= \Gamma\big(\lambda_n+a'(n)f(x^*(\lambda_n))+a'(n)\xi_n\big)-\beta.
\end{gather}
Since the map $\lambda \to x^*(\lambda)$ is concave decreasing and piecewise linear, we have for some finite constant $0<L_5\leq L_6$ such that
\begin{gather*}
  -L_6(\lambda_1-\lambda_2)\leq f(x^*(\lambda_1))-f(x^*(\lambda_2))\leq  -L_5(\lambda_1-\lambda_2)
\end{gather*}
Replace $\lambda_1$ by $\lambda_n$ and $\lambda_2$ by $\beta$. Since $f(x^*(\beta))=0$:
\begin{gather*}
     -L_6(\lambda_n-\beta)\leq f(x^*(\lambda_n))\leq -L_5(\lambda_n-\beta)
\end{gather*}
Thus,
\begin{gather}
    \nonumber \Gamma\big(\lambda_n-a'(n)L_6(\lambda_n-\beta)+a'(n)\xi_n\big)-\beta\leq  \lambda_{n+1}-\beta \leq \\ \nonumber \Gamma\big(\lambda_n-a'(n)L_5(\lambda_n-\beta)+a'(n)\xi_n\big)-\beta \nonumber\\
    \nonumber |\lambda_{n+1}-\beta| \leq \nonumber \max(|\Gamma\big(\lambda_n-a'(n)L_6(\lambda_n-\beta)+a'(n)\xi_n\big)\\-\beta|,  |\Gamma\big(\lambda_n-a'(n)L_5(\lambda_n-\beta)+a'(n)\xi_n\big)-\beta|)
\end{gather}
Using the fact that $\Gamma(\beta)=\beta$ along with the non-expansive property of $\Gamma(\cdot)$, we have:
\begin{gather}
    \nonumber |\lambda_{n+1}-\beta| \leq \max(|(1-a'(n)L_6)(\lambda_n-\beta)+a'(n)\xi_n|, \\ \nonumber |(1-a'(n))L_5(\lambda_n-\beta)+a'(n)\xi_n|
\end{gather}
For some $\hat{n}\geq N$ such that $a'(n)L_6<1 \ \forall \ n\geq \hat{n}$,
\begin{gather*}
    |\lambda_{n+1}-\beta|\leq \max(|(1-a'(n)L_6)|\lambda_n-\beta|+a'(n)|\xi_n|, \\  |(1-a'(n)L_5)|\lambda_n-\beta|+a'(n)|\xi_n|)\\
    \leq (1-a'(n)L_5)|\lambda_n-\beta|+a'(n)|\xi_n |, \ n \geq \hat{n}.
\end{gather*}
Iterating the above inequality for $\hat{n}\leq m\leq n$, we get:
\begin{gather}
    \nonumber |\lambda_{n+1}-\beta|\leq |\lambda_{\hat{n}}-\beta|\prod_{m=\hat{n}}^n(1-a'(m)L_5))\\ +\sum_{m=\hat{n}}^n\big(\prod_{k=m+1}^n(1-L_5a'(k))\big)a'(m)|\xi_m|.
\end{gather}
Notice that
$$|\xi_m|=|f(x_m)-f(x^*(\lambda_m))|\leq L_4\|x_m-x^*(\lambda_m)\|.$$ 
Using this and the first part of the proof we can bound each $|\xi_m|$ with a term similar to r.h.s.\ of Theorem \ref{shaan_th}(a). Using the same analysis as used for deriving (\ref{psi_bound}), one can verify that 
$$\sum_{m=\hat{n}}^n\big(\prod_{k=m+1}^n(1-L_5a'(k))\big)a'(m)\leq \frac{1}{L_5}.$$ 
This leads us to
\begin{gather}
    \nonumber |\lambda_{n+1}-\beta|\leq |\lambda_{\hat{n}}-\beta|e^{-(1-\alpha)L_5b'_{\hat{n}}(n)}\\ +
    \frac{L_4}{L_5}\Big(e^{-(1-\alpha)b_{\hat{n}}(n)}\|x_{\hat{n}}-x^*(\lambda_{\hat{n}})\|+\frac{\delta+a(\hat{n})c_1+\bar{\epsilon}(\hat{n})c_3}{1-\alpha}\Big)
\end{gather}
holds with probability exceeding
    \begin{gather}
    1-2d\sum_{m\geq\hat{n}+1}^ne^{-D\delta^2/\beta_{\hat{n}}(m)}, 0<\delta\leq c,\\
    1-2d\sum_{m\geq\hat{n}+1}^ne^{-D\delta/\beta_{\hat{n}}(m)}, \delta> c
    \end{gather}
where we applied union bound to get the probability. This completes the proof of Theorem \ref{shaan_th}(b).

\section{Comparison with RVI Q-learning and Simulation Results}
\begin{figure}
\centering
\begin{subfigure}{0.45\textwidth}
    \includegraphics[width=\textwidth]{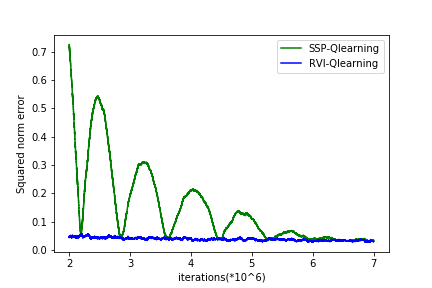}
    \caption{20 states, 5 actions.}
    \label{fig:first}
\end{subfigure}
\hfill
\begin{subfigure}{0.45\textwidth}
    \includegraphics[width=\textwidth]{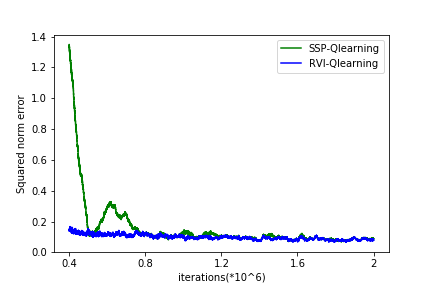}
    \caption{20 states, 5 actions.}
    \label{fig:second}
\end{subfigure}
\caption{Squared Error vs number of iterations}
\label{fig:figures}
\end{figure}

\subsection{RVI Q-learning}
The algorithm for RVI Q-learning is given by \cite{abounadi}
\begin{eqnarray}\label{RVI}
    Q_{n+1}(i,u)&=&k(i,u)+
\sum_{j\in \mathcal{S}}p(j|i,u)\min_vQ_n(j,v) \nonumber \\&&-Q_n(i_0,u_0)
\end{eqnarray}
with $i_0\in S$, $u_0\in A$ prescribed. The purpose of subtracting the scalar “offset” $Q_n(i_0,u_0)$ from each component on the r.h.s. of (\ref{RVI}) is to keep the iterations stable—recall that $Q(\cdot,\cdot)$ is specified anyway only up to an additive constant. It turns out that $Q(i_0,u_0)\to \beta$. 
The asynchronous RVI Q-learning algorithm is
\begin{gather}\label{a_stoc_RVI}
    \nonumber Q_{n+1}(i,u)=Q_n(i,u)+a(n)I\{X_n=i,Z_n=u\}\big(k(i,u)\\ +\min_vQ_n(X_{n+1},v)-Q_n(i_0,u_0)-Q_n(i,u)\big).
\end{gather}
Under some additional conditions such as aperiodicity and comparably frequent updates of all components, the above converges to a unique $Q(\cdot,\cdot)$ with $Q(i_0,u_0)=\beta$ \cite{abounadi}.
\subsection{Simulation experiments}
We took two controlled Markov chains with 20 states and 5 actions in each state as examples. Fig. \ref{fig:first} corresponds to the first, while Fig. \ref{fig:second} corresponds to the second. We wanted Markov chains which are irreducible under any policy and the easiest way to generate such a chain is to generate a random transition matrix with each element drawn randomly from the interval $[0,1]$ with a high probability, and then  normalize the rows. For the second example, after getting the random matrix, we changed 50\% of the entries to $0$ except the entries corresponding to $p(i|0,u)$ and $p(0|i,u)$ for all $i\in \mathcal{S}$, $u\in \mathcal{U}$, and then normalized the rows. This yielded an irreducible matrix for the two examples.  For running both algorithms, we used the same step size of $a(n)=\frac{1}{\lceil{\frac{n}{2}\rceil}^{0.65}}$ and $a'(1.5|\mathcal{S}||\mathcal{U}|n)=\frac{1}{\lceil{\frac{5000 + n}{1.5|\mathcal{S}||\mathcal{U}|}\rceil}^{0.65}\log(\lceil{\frac{5000 + n}{1.5|\mathcal{S}||\mathcal{U}|}\rceil})}$, with $\lambda_n$ not being updated for $n$ that are not multiples of $1.5|\mathcal{S}||\mathcal{U}|$. The reason for choosing these step sizes was relatively fast rate of convergence for both algorithms and less noisy and clear plots.

Since we had the access to the transition matrices, we first obtained the actual value of the desired $Q^*$ for each algorithm using value iteration. Then we ran the asynchronous version of both algorithms and stored the $l_2$-norm error relative to this $Q^*$. While explicit contractiveness of the SSP scheme with respect to a certain norm  (as opposed to non-expansiveness of RVI) might lead us to expect better convergence for SSP, the preliminary experiments reported here indicate otherwise. RVI was quick to converge, whereas SSP had initially oscillatory behaviour which died down after some time and it caught up with RVI. This appears to be due to the high sensitivity of $f(Q_n)$ to $\lambda$ which is updated on a slower time scale. However, this effect is significantly subdued in the second example. These and another set of similar examples (not reported here) suggest that if the graph of the MDP is well-connected, suggesting rapid mixing under a stationary policy, it favours RVI over SSP and this advantage diminishes as the connectivity becomes sparser. This, however, needs more extensive experiments and theoretical study. 

A further future research direction is to combine either scheme with suitable function approximation.
\bibliographystyle{IEEEtran}
\bibliography{references}

\begin{thebibliography}{1}
\providecommand{\url}[1]{#1}
\csname url@rmstyle\endcsname
\providecommand{\newblock}{\relax}
\providecommand{\bibinfo}[2]{#2}
\providecommand\BIBentrySTDinterwordspacing{\spaceskip=0pt\relax}
\providecommand\BIBentryALTinterwordstretchfactor{4}
\providecommand\BIBentryALTinterwordspacing{\spaceskip=\fontdimen2\font plus
\BIBentryALTinterwordstretchfactor\fontdimen3\font minus
  \fontdimen4\font\relax}
\providecommand\BIBforeignlanguage[2]{{%
\expandafter\ifx\csname l@#1\endcsname\relax
\typeout{** WARNING: IEEEtran.bst: No hyphenation pattern has been}%
\typeout{** loaded for the language `#1'. Using the pattern for}%
\typeout{** the default language instead.}%
\else
\language=\csname l@#1\endcsname
\fi
#2}}

\bibitem{Watkins}
\BIBentryALTinterwordspacing
C.~J. C.~H. Watkins, ``Learning from delayed rewards,'' Ph.D. dissertation,
  King's College, Cambridge, UK, May 1989. [Online]. Available:
  \url{http://www.cs.rhul.ac.uk/~chrisw/new_thesis.pdf}
\BIBentrySTDinterwordspacing

\bibitem{abounadi}
J.~Abounadi, D.~Bertsekas, and V.~S. Borkar, ``Learning algorithms for markov
  decision processes with average cost,'' \emph{SIAM Journal on Control and
  Optimization}, pp. 681--698, 2001.

\bibitem{Puterman}
M.~L. Puterman, \emph{Markov decision processes: discrete stochastic dynamic
  programming}.\hskip 1em plus 0.5em minus 0.4em\relax John Wiley \& Sons,
  2014.

\bibitem{bert}
\BIBentryALTinterwordspacing
D.~P. Bertsekas, ``A new value iteration method for the average cost dynamic
  programming problem,'' \emph{SIAM J. Control Optim.}, vol.~36, no.~2, p.
  742–759, mar 1998. [Online]. Available:
  \url{https://doi.org/10.1137/S0363012995291609}
\BIBentrySTDinterwordspacing

\bibitem{chandak}
\BIBentryALTinterwordspacing
V.~S.~B. Siddharth~Chandak and P.~Dodhia, ``Concentration of contractive
  stochastic approximation and reinforcement learning,'' \emph{CoRR}, vol.
  abs/2106.14308, 2021. [Online]. Available:
  \url{https://arxiv.org/abs/2106.14308}
\BIBentrySTDinterwordspacing

\bibitem{Zhang}
\BIBentryALTinterwordspacing
S.~Zhang, Z.~Zhang, and S.~T. Maguluri, ``Finite sample analysis of
  average-reward td learning and q-learning,'' in \emph{Advances in Neural
  Information Processing Systems}, M.~Ranzato, A.~Beygelzimer, Y.~Dauphin,
  P.~Liang, and J.~W. Vaughan, Eds., vol.~34.\hskip 1em plus 0.5em minus
  0.4em\relax Curran Associates, Inc., 2021, pp. 1230--1242. [Online].
  Available:
  \url{https://proceedings.neurips.cc/paper/2021/file/096ffc299200f51751b08da6d865ae95-Paper.pdf}
\BIBentrySTDinterwordspacing

\bibitem{tsitsiklis}
\BIBentryALTinterwordspacing
J.~Tsitsiklis, ``{A}synchronous stochastic approximation and {Q}-learning,''
  LIDS Res.\ Cent., MIT, Tech. Rep., 1993. [Online]. Available:
  \url{http://www.springerlink.com/index/QX335286076X3566.pdf}
\BIBentrySTDinterwordspacing

\end{thebibliography}

\end{document}